\documentclass[
]{ceurart}

\usepackage{listings}
\usepackage{xcolor}
\usepackage{threeparttable}
\begin{document}

\copyrightyear{2022}
\copyrightclause{Copyright for this paper by its authors.
  Use permitted under Creative Commons License Attribution 4.0
  International (CC BY 4.0).}

\conference{Sci-K'25: International Workshop on Scientific Knowledge: Representation, Discovery, and Assessment,
  November 2-6, 2025, Nara, Japan}

\title{What Are Research Hypotheses?}


\author[1]{Jian Wu}[%
orcid=0000-0003-0173-4463,
email=j1wu@odu.edu,
url=https://www.cs.odu.edu/~jwu/,
]
\cormark[1]
\address[1]{Old Dominion University}

\author[2]{Sarah Rajtmajer}[%
orcid=0000-0002-1464-0848,
email=smr48@psu.edu,
url=https://www.rajtmajerlab.net/,
]
\cormark[1]
\address[2]{The Pennsylvania State University}


\cortext[1]{The two authors made equal contributions to this paper.}

\begin{abstract}
Over the past decades, alongside advancements in natural language processing, significant attention has been paid to training models to automatically extract, understand, test, and generate hypotheses in open and scientific domains. However, interpretations of the term \emph{hypothesis} for various natural language understanding (NLU) tasks have migrated from traditional definitions in the natural, social, and formal sciences. Even within NLU, we observe differences defining hypotheses across literature. In this paper, we overview and delineate various definitions of hypothesis. Especially, we discern the nuances of definitions across recently published NLU tasks. We highlight the importance of well-structured and well-defined hypotheses, particularly as we move toward a machine-interpretable scholarly record.
\end{abstract}
\begin{keywords}
natural language processing,
natural language understanding,
natural language inference,
hypothesis extraction,
hypothesis generation,
scientific claim verification,
large language models\end{keywords}

\maketitle

\section{Introduction}

The word ``hypothesis'' has been used variably with different meanings, over decades and centuries, across the social, natural, and formal sciences---from its conceptual roots in ancient Greek philosophy \cite{kuhn1963structure,popper1963science,lakatos1970history} to the development of hypothesis testing as statistical methods \cite{fisher1928statistical,fisher1955statistical} and subsequent evolution of the term in different fields reflecting their unique questions and approaches \cite{glass2008brief}. Of course, ambiguity around language and variable use of terminology is pervasive within and outside of science. Language has always been an impoverished tool for representation and expression of complex ideas \cite{fauconnier1997mappings,malt2013thought,yarkoni2022generalizability}. In many cases though, this is not a problem. Members of a particular community develop shared understanding of the meaning of a given term in context, and this allows them to communicate effectively toward collective goals. 

We argue that ambiguity and variability around the definition of hypotheses, which was once acceptable---even perhaps productive---is now a critical concern in light of natural language processing (NLP) and natural language understanding (NLU) tasks requiring quantitative operationalization of hypotheses, in particular, hypothesis extraction (detection/identification), verification, and generation. Emerging technical work in these fields often do not include explicit definitions of hypotheses, claims, or evidence, instead relying \emph{de facto} on benchmark datasets to provide implicit definitions.

Following, we survey the definitions and operationalizations of hypotheses, focusing on research hypotheses engaged in the hypothesis mining literature. Research hypotheses are hypotheses designed for systematic investigation within a research framework. In this paper, we do not distinguish between research hypothesis and scientific hypothesis. In principle, these two terms have different scopes, but in practice, they are often used interchangeably in modern hypothesis mining papers. The related tasks are particularly in the areas of natural language inference (NLI), hypothesis extraction, scientific hypothesis evidencing and scientific claim verification, and scientific hypothesis generation. We highlight important differences and discuss the challenges these differences impose on knowledge assembly and aggregation. 
Our work is motivated by the vision of a computable scholarly record---a verifiable and extensible knowledge base synthesizing computationally and data-enabled discoveries \cite{stodden2024emergent}. This vision, we suggest, will be enabled by machine-readable hypotheses well-structured in predictable formats. 



\section{Background}

\subsection{Conceptual origins} 

The word \emph{hypothesis} derives from Greek and means, literally, \emph{a putting under} or supposition. Ancient Greek philosophers used the term to describe a foundational assumption upon which to build out further reasoning. Plato uses the term in several of his dialogues with this intention--namely, as a claim accepted temporarily in order to explore its implications. Of particular interest to Plato was whether a hypothesis could support consistent and coherent conclusions \cite{karasmanes1987hypothetical,ausland2022socrates}. Aristotle also engaged with the term. Aristotle viewed hypotheses more cautiously, being skeptical of relying on hypotheses without empirical verification. He delineated tentative assumptions from axioms or first principles, insisting that scientific knowledge must be based on demonstrated causes, not just assumed premises \cite{barnes1994posterior}. 

During the scientific revolution of the 16th and 17th centuries, the concept of hypothesis evolved significantly. Galileo and Newton began using hypotheses as formal components of the scientific method, emphasizing the importance of testing through observation and experiment \cite{galilei1914dialogue,newton1833philosophiae}. René Descartes also contributed to this shift, promoting skepticism and the formulation of testable propositions \cite{descartes1912discourse,sakellariadis1982descartes}.

By the 19th and 20th centuries, the hypothesis had become a central pillar in science. Prominent philosophers of science Karl Popper and Thomas Kuhn centered hypotheses within the scientific process, both agreeing that hypotheses must engage with empirical data in some way, i.e., should be testable, observable \cite{quinn1983hypothesis,leplin1997novel}.  However, Kuhn and Popper's views on scientific advancement differed in important ways and their respective views on the role of hypotheses reflected these differences. Kuhn, a historian, viewed periods of science through the lens of \emph{paradigms} and hypotheses as statements that operates within a paradigm (vs. free-floating assumptions to be directly tested in isolation) \cite{shapere1964structure}. Popper, on the other hand, highlighted the asymmetry between verification and \emph{falsification}—hypotheses cannot be proven true, only proven false. 
Central to Popper's thinking is his assertion that confirmations for a theory are easy to find if we look for them. Confirming evidence should only count when it is the result of a genuine test of the theory (i.e., we conclude that the theory withstood an attempt to disconfirm it) \cite{lakatos1970history}. In essence, Popper argued that Kuhn's hypotheses risked self-fulfillment. Popperian theories underlie the current open science movement triggered by the replication crisis, including efforts promoting development of strong, \emph{testable} hypotheses, and preregistration to delineate exploratory vs. confirmatory findings \cite{rajtmajer2022failure,nosek2018preregistration}.

\subsection{Modern forms}
In the last decades, there has been a growing interest in hypothesis mining in scientific literature, mostly in the fields of NLP and NLU, but also in interdisciplinary fields between social science and AI. The exact definitions of hypotheses involved are not always provided in the context of research problems, and the specific forms and expressions vary across papers. Here, we categorize modern hypotheses into several types, which may deviate from traditional definitions, e.g., \cite{mueller2019deepcause}. 

\textbf{Ideas as hypotheses.} \label{sec:ideas}
As defined in Kuhn \& Hawkins \cite{kuhn1963structure}, an \emph{idea} is a realization or hypothesis that can challenge and shift paradigms within a scientific community. 
Several recent papers about hypothesis generation adopted this conceptualization and treated ideas as hypotheses \cite{kumar2024can,wang2024scimon}. In Kumar et al. \cite{kumar2024can}, the authors build a dataset containing \emph{Future Research Ideas (FRIs)} and then consider the generated FRIs to be hypotheses.
The structure of these ideas includes: premises; a traditional research hypothesis; and its context. In Wang et al. \cite{wang2024scimon}, the authors do not discern the term ideas from hypotheses and use them interchangeably in certain contexts, but the ground truth data shows that the generated content contains preliminary and broad notations intended to inspire further investigation, which is aligned with the concept of ideas. An example is shown in Table~\ref{tab:examples}.  

\textbf{Claims as hypotheses.} \label{sec:hypothesesasclaims}
The classical definition of \emph{claim} is the conclusion or assertion that you want your audience to accept \cite{toulmin2003uses}. Adopting this definition, in scientific literature, a claim can be defined as a specific assertion reported as a finding of the paper. A paper can make more than one claim, and a claim may contain one or multiple sentences. One definition of hypothesis is a claim that has not been tested \cite{heger2024natural}. In Alipourfard et al. \cite{alipourfard2021systematizing}, authors label a \emph{claim trace} for each paper in their corpora, and each trace contains four claims. Hypotheses and evidence are treated as two types of claims. In the recent SciHyp dataset \cite{vasu2024scihyp} developed for hypothesis detection and classification in Computer Science papers, many hypotheses in the ground truth are claims manually extracted throughout the full text. This ambiguity also occurs in scientific hypothesis evidencing (SHE; \cite{koneru2024coling}) and scientific claim verification (SCV; \cite{wadden2020scifact}) tasks. Both tasks aim to discern the relationship (or stance) between a hypothesis (in SHE, or a claim in SCV) and a candidate piece of evidence. 

\textbf{Hypothesis-proposals.}\label{sec:proposals}
In recent works about hypothesis generation, models are built to generate not only a hypothesis but also a series of related sections such as its background, justification, and test procedures, resulting in a \emph{proposal}-style document \cite{gottweis2025aiscientist}. The hypothesis-proposal increases the transparency of hypothesis generation and provides a guide for testing. However, the specific format/sections of the proposal differ by model. An example in \cite{gottweis2025aiscientist} is shown in Table~\ref{tab:examples}. 

\textbf{Formal expressions.}
In early work, a research hypothesis is broken down into three dimensions, namely contexts, variables, and relationships \cite{descartes1912discourse}. Each hypothesis is associated with a target variable and a set of independent variables, and relationships refer to the interactions between a given set of variables under a given context that produces the hypothesis. A hypothesis is then naturally expressed with a \emph{semantic tree} in which the nodes represent variables and the edges represent relationships. 

In more recent works in NLU, papers have expressed research hypotheses in various ways, depending on the focal tasks. 
For example, in hypothesis generation tasks, the generated hypothesis may be composed of multiple declarative statements, in which one serves as the main hypothesis and the others provide additional context or details (see\cite{ghafarollahi2025sciagents} and \cite{wang2024scimon} in Table~\ref{tab:examples}). In the hypothesis evidencing task, hypotheses can be written as questions \cite{koneru2024coling}, which can be converted into hypotheses in declarative form. A research hypothesis can also be decomposed as a question and an answer, e.g., the SciTail dataset \cite{khot2018scitail}. Their entailment relation can be further explained using an \emph{entailment tree} showing how the hypothesis follows from the text corpus \cite{dalvi2021entailmentbank} (Figure~\ref{fig:entailmenttree}). 

\section{Scientific hypothesis-related tasks and datasets}

\subsection{Natural language inference}
Natural language inference (NLI, i.e., recognizing textual entailment (RTE)) involves assessing whether a given textual premise entails or implies a given hypothesis \cite{dagan2022recognizing,storks2019recent}. Most NLI datasets, such as SNLI \cite{do2020snli} and RTE-6 \cite{bentivogli2009rte6}, are in open domains \cite{do2020snli}. SciTail is one of the few datasets built for scientific NLI \cite{khot2018scitail}. 
Hypotheses are expressed in single declarative sentences (see Table~\ref{tab:examples}). 

\begin{figure}
    \centering
    \includegraphics[width=.9\textwidth]{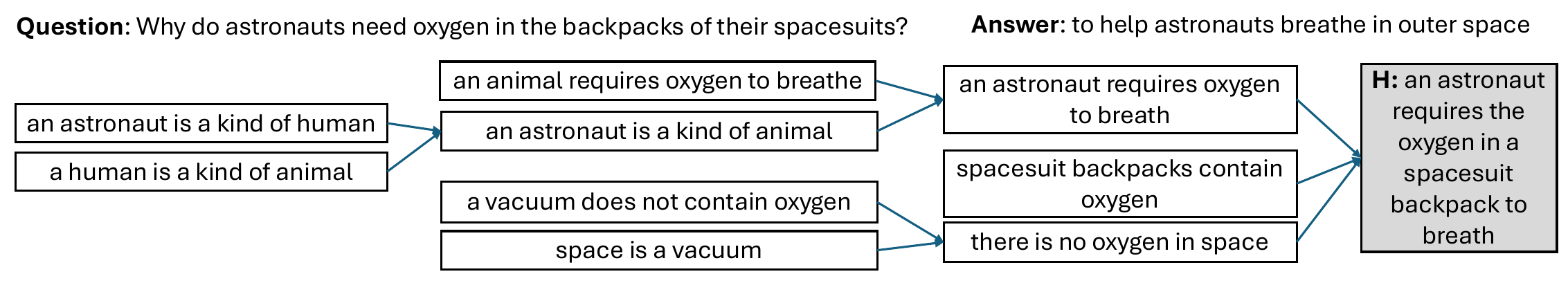}
    \caption{An example in the {\sf EntailmentBank} dataset, demonstrating the entailment tree structure to support the hypothesis (in a shaded box) generated based on the question and answer. The figure is adopted from \cite{dalvi2021entailmentbank}.}
    \label{fig:entailmenttree}
\end{figure}
Another scientific NLI dataset is {\sf EntailmentBank}, built for multi-step scientific inference (Figure~\ref{fig:entailmenttree}). The task is to generate an entailment tree given a hypothesis. The tree shows a hierarchical supportive structure of claims toward a hypothesis. Other scientific NLI datasets include MediNLI \cite{shivade2017mednli} and BioNLI \cite{bastan2022bionli} in the medical and biomedical domains, and e-SNLI-VE \cite{do2020snli} for visual scientific inference. 

\subsection{Hypothesis and claim extraction} 
Here, the goal is to automatically identify hypotheses from a scientific document. Because hypotheses can be viewed as claims prior to testing, the input, output, and methods for extracting hypotheses and claims are similar. 
The input document can be an abstract, e.g., \cite{achakulvisut2019claim,wei2023claimdistiller} or a full paper \cite{alipourfard2021systematizing}. In White et al. \cite{white2011cisp}, authors propose and apply a schema 
for annotating sentences in full text of scientific articles into 9 types: hypothesis; goal; motivation; background; method; experiment; result; observation; and conclusion \cite{white2011cisp}. In the dataset used for training DeepCause, a model for hypothesis extraction, selected claims identified from the full text are labeled as hypotheses \cite{mueller2019deepcause}. 

Claim extraction may benefit from structured abstracts which contain, e.g., a dedicated \emph{Findings} section (see Lancet \cite{lancet}). Yet still, the specific sections differ across journals. For example, in \cite{achakulvisut2019claim}, 
\emph{findings} or \emph{proposed items} are labeled as claims and one abstract may contain multiple claims (Table~\ref{tab:examples}). 

It is worth noting that the papers of computer science and several other domains often state findings or contributions without explicitly stating hypotheses, e.g., \cite{wang2024scimon}. We may call these findings or contributions \emph{pseudo-hypotheses} because perhaps the authors started with them as hypotheses and empirically tested them to be valid. 

\subsection{Scientific hypothesis evidencing and scientific claim verification}
\begin{figure}[t]
    \centering
    \includegraphics[width=\textwidth]{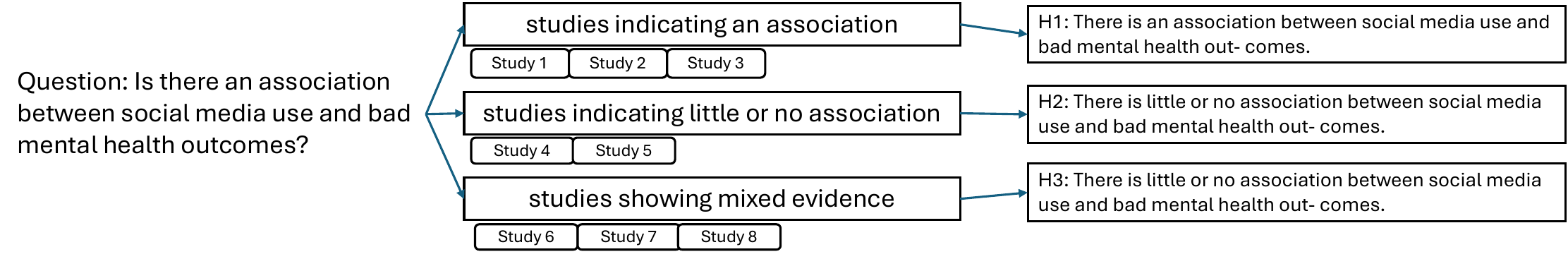}
    \caption{The collaborative review document structure for an example question and the hypotheses derived from the (question, answer) pairs. \cite{koneru2024coling}. }
    \label{fig:she}
\end{figure}

Scientific hypothesis evidencing (SHE) \cite{koneru2024coling} is the task of automatically identifying evidence from scientific publications in support of or refutation of a given hypothesis. This task is similar to another task called Scientific claim verification (SCV), both reflecting a model's reasoning capability. The main difference is that in SHE, the hypotheses are usually high-level research questions (Figure~\ref{fig:she}). In SCV datasets, the hypotheses are usually lower-level claims in specific contexts. However, certain cases are in between. Both tasks can be divided into two subtasks: identifying evidence candidates from a corpus of documents; and discerning the relationship between the hypothesis (claim) and an evidence candidate. 
Most research focuses on the second subtask, in which the relationships are classified into exclusive categories, namely \emph{SUPPORT}, \emph{REFUTE}, and \emph{NEI} (not enough information) or their variants. 

In Koneru et al. \cite{koneru2024coling} authors build a dataset for the task of SHE using community-driven annotations of studies in social sciences. 
The input is a hypothesis and an abstract (i.e., candidate evidence), and the output is a label indicating whether the abstract entails, contradicts, or is inconclusive to the hypothesis. In Wadden et al. \cite{wadden2020scifact}, authors build an SCV dataset, the input of which is a (claim, abstract) pair (see an example in Table~\ref{tab:examples}). 
In the Covid-Fact dataset \cite{saakyan2021covid}, claims are obtained by filtering titles of social media posts. While, in the HealthVer dataset, claims are manually extracted from questions and snippets returned by search engines \cite{sarrouti2021healthver}.

DiscoveryBench \cite{majumder2025discoverybench} is a benchmark designed for a task called \emph{data-driven discovery}. Similar to SCV, the goal is to verify a hypothesis, originally expressed in the form of a research question.
Nevertheless, instead of using abstracts as evidence, the verification is grounded in data. In an example task, a model is given a dataset in the form of a spreadsheet and a research question. The model is expected to generate a stepwise workflow that tries to answer the question using the given data. The final output, treated as a hypothesis, is decomposed into a semantic tree containing (context, variable, relationship) and compared against the gold standard hypothesis (see an example in Table~\ref{tab:examples}). 
{\sf SciClaimHunt} is a dataset recently built for SCV, in which a small amount of claims are manually extracted from the discussion and conclusion sections of research papers in computer science. Most claims are generated by LLMs. Of note, a fraction of claims are not self-contained and require reference to the context of the source paper. 

\begin{table}
\caption{Examples of input and output for selected hypothesis-related tasks. Task names are abbreviated. Gen=Hypothesis generation; Ext=Hypothesis (Claim) extraction; SCV=Scientific claim verification; DDD=Data-driven discovery.}\label{tab:examples}\footnotesize
\begin{tabular}{p{1cm}|p{2.1cm}|p{2.0cm}|p{9.5cm}}
\toprule
\textbf{Tasks} & \textbf{Input}     &  \textbf{Output} & \textbf{Examples}\\
\midrule
\multirow{5}{*}{NLI} & \multirow{5}{*}{\parbox{2.1cm}{(Premise,\\Hypothesis)\\ (SciTail \cite{khot2018scitail})}}&\multirow{5}{*}{\parbox{2.0cm}{Label:\\ Entail,\\Neural,\\Contradiction}} & Input premise: Beats are the periodic and repeating fluctuations heard in the intensity of a sound when two sound waves of very similar frequencies interfere with one another.\\
      & & & \emph{Input Hypothesis: When waves of two different frequencies interfere, beating occurs.}\\
      & & & \emph{Output label: entail}\\
\midrule
\multirow{5}{*}{Ext}&\multirow{5}{*}{\parbox{2.1cm}{Full text or\\an abstract\\ \cite{achakulvisut2019claim}}} & \multirow{5}{*}{\parbox{2.0cm}{Hypotheses or\\claims}} & Input: The abstract of a paper titled \emph{A Morphological Hessian Based Approach for Retinal Blood Vessels Segmentation and Denoising Using Region Based Otsu Thresholding} \cite{bahadarkhan2016plosone}\\
        &&& \emph{Output claim: We proposed a less computational unsupervised automated technique with promising results for detection of retinal vasculature by using morphological hessian based approach and region based Otsu thresholding.}\\
\midrule
\multirow{5}{*}{SCV}&\multirow{5}{*}{\parbox{2.1cm}{(claim,\\abstract)\\ (SciFact \cite{wadden2020scifact})}} & \multirow{5}{*}{\parbox{2.0cm}{Label:\\ Support,\\Refute,\\NEI}} & Input claim: The coronavirus cannot thrive in warmer climates.\\
      & & & Input abstract: ...most outbreaks display a pattern of clustering in relatively cool and dry areas...This is because the environment
can mediate human-to-human transmission of SARS-CoV-2, and unsuitable climates can cause the virus to destabilize quickly...\\
      & & & \emph{Output label: Support}\\
\midrule
\multirow{5}{*}{DDD}&\multirow{5}{*}{\parbox{2.1cm}{Goal and\\data\\ (DiscoveryBench \cite{majumder2025discoverybench})}} & \multirow{5}{*}{\parbox{2.0cm}{Workflow\\answer\\and\\ decomposed\\hypothesis}} & Input goal: How did urban land use affect the invasion of different types of introduced plants in Catalonia?\\
    & & & Input data: a relational table of habitat type vs. plant type\\
    & & & \emph{Output workflow: A workflow to answer the question.}\\
    & & & \emph{Output answer (hypothesis): Urban land use increased
invasion by agriforest plants over gardening introduced ones in Catalonia.}\\
    & & & \emph{Output (context, variable, relationship)=(urban habitat type, gardening, unintentional, reduced) }\\
\midrule
\multirow{9}{*}{Gen}& \multirow{9}{*}{\parbox{2.1cm}{Keywords\\(SciAgents \cite{ghafarollahi2025sciagents})}} & \multirow{9}{*}{Proposal} & Input keywords: heat transfer performance, soft lithography, etc.\\ 
       & & & \emph{Output hypothesis: We hypothesize that integrating biomimetic materials with microfluidic chips will significantly enhance their heat transfer performance and biocompatibility, making them ideal for advanced biomedical applications. Specifically, we propose that the lamellar structure of biomaterials, inspired by keratin scales, can be engineered into microfluidic chips using soft lithography techniques to improve their mechanical behavior and heat transfer efficiency under cyclic loading conditions. } \\
       & & & \emph{Output other sections: outcome, mechanisms, design principles, unexpected properties, comparison, and novelty, and their expanded versions.}\\
        \midrule
\multirow{3}{*}{Gen}&\multirow{3}{*}{Data \cite{zhou2024hypothesis}}  & \multirow{3}{*}{Hypothesis} & Input: Tweet pairs in the {\sc Tweet Popularity} dataset \cite{tan2014effect}.\\
        & & & \emph{Output: Tweets with named entities like people, places, or organizations tend to get more retweets by being more specific.}\\
\midrule
\multirow{9}{*}{Gen}&\multirow{8}{*}{\parbox{2.1cm}{Background (SciMon \cite{wang2024scimon})}}   & \multirow{8}{*}{Ideas} & Input seed term: diverse relational edge embedding\\
       & & & Input background: the task of converting a natural lanuage question into an executable sql query , known as text - to -sql, is an important branch of semantic parsing . the state - of - the -art graph -based encoder has been successfully used in this task but does not model the question syntax well. \\
        && & \emph{Output: We propose a novel graph-based encoder that uses a diverse relational edge embeddings to model the question syntax. }\\
\bottomrule
\end{tabular}

\end{table}

\subsection{Scientific hypothesis generation}
The goal of scientific hypothesis generation is to automatically create new, testable scientific hypotheses or research ideas that identify novel relationships, phenomena, or gaps in existing knowledge \cite{kumar2024can}. Recent advancements in LLMs, e.g., Llama \cite{touvron2023llama2} and GPT \cite{brown2020gpt2}, offer promise. Here too, existing literature is inconsistent with respect to task formulation, i.e., \emph{input} and \emph{output} (see examples in Table~\ref{tab:examples}). 

We identify four types of input for the hypothesis generation tasks:

\noindent(1) \emph{keywords}, concise descriptions of the topics or central concepts of the model, such as in SciMon \cite{wang2024scimon} and SciAgents \cite{ghafarollahi2025sciagents}; 

\noindent (2) \emph{goals}, brief discourse outlining research goals. In AI co-scientist \cite{gottweis2025aiscientist}, 
goals can be a request to propose a novel hypothesis, suggest special requirements, or ask a question. In Si et al. \cite{si2025iclr}, LLMs are provided ``topics'', such as ``novel prompting methods that can better quantify uncertainty or calibrate the confidence of large language models'', which serve as goals. In Pu et al. \cite{pu2025piflow}, an input is described as an ``objective'', which is equivalent to a goal;

\noindent (3) \emph{data}, i.e., a dataset, based on which the model is requested to generate a hypothesis, e.g., \cite{zhou2024hypothesis}; 

\noindent (4) \emph{background}, context, rationale, or theoretical foundation of a hypothesis. For example, in the SciMon framework \cite{wang2024scimon}, the input includes seed terms, including concepts and keywords, and background context, which contains problems, motivations, or focus points. The Mamba framework \cite{chai2024mamba} uses the same ground truth as SciMon \cite{wang2024scimon}. The MOOSE framework \cite{yang2024moose} uses background and inspiration derived from the raw web corpus as the input. 

Likewise, we identify three types of output of hypothesis generation tasks: 

\noindent(1) \emph{traditional hypothesis}, usually expressed as a single or multiple declarative sentences, e.g., \cite{zhou2024hypothesis,qi2023large}. 

\noindent (2) \emph{ideas}, enriched hypotheses as shown in Section~\ref{sec:ideas}. For example, in {\sc SciMon}, generated ideas may contain claims, methods, and objectives extracted from abstracts. 

\noindent (3) \emph{hypothesis proposals}, comprehensive structured hypothesis description documents. For example, the output of SciAgents \cite{ghafarollahi2025sciagents} is a document containing hypotheses, outcomes, mechanisms, design principles, unexpected properties, comparison, and novelty, each having its expanded version. AI co-scientist \cite{gottweis2025aiscientist} also outputs a structured document but with different sections: introduction, recent findings, related research, rationale, specificity, experimental design, and validation. 
Si et al. \cite{si2025iclr} request LLM agents to generate an ``idea'', containing several components (e.g., problem, existing methods, motivation, proposed method, experiment plan), similar to a research proposal. Whereas, the Piflow framework \cite{pu2025piflow} requires LLM agents to generate a ``hypothesis structure" consisting of rationale, hypothesis, reiterate, and an experimental candidate. 


\section{Discussion and Conclusions}

Our work here intends to be more descriptive than prescriptive. We outline the various definitions and instances of hypotheses in existing scientific literature (and beyond). In particular, we focus on definitions of hypotheses and related concepts in recent work in NLU. 

We hope that this work may raise awareness in the communities of hypothesis mining communities about standardizing some of to support corpus-level tools, e.g., knowledge graphs representing connections amongst interdisciplinary hypotheses, or hypothesis generation models across multiple domains. In lieu of standardization, the inclusion of explicit, clear definitions of hypotheses (formal, where possible) could improve alignment and assembly. 

For more than two decades, many in the research community have advocated for open data, open materials, preregistration, and other best practices as central to the vision of a \emph{searchable and interpretable scholarly record}.  With recent technological advances, this vision is on the horizon. The ultimate goals of an interpretable scholarly record are: robust and efficient scientific progress; thoughtful allocation of community resources toward important open problems; and honest dialogue with public and policymakers.
How--precisely--a queryable scholarly corpus comes together is an open question. 
Here, we suggest that dissemination of clear, consistent, well-specified machine-readable hypotheses, claims, and evidence are critical to this mission.








\bibliography{ref}



\end{document}